\documentclass[letterpaper, 10 pt, journal, twoside]{IEEEtran}  
\IEEEoverridecommandlockouts                              

\usepackage[nolist]{acronym} 
\usepackage{amsfonts,siunitx,booktabs,cite} 
\usepackage{graphics} 
\usepackage{epsfig} 
\usepackage{times} 
\usepackage{amsmath} 
\usepackage{amssymb}  
\usepackage[linesnumbered,ruled]{algorithm2e}
\usepackage{stmaryrd}
\usepackage{soul}
\usepackage{bm}

\usepackage{amsthm}

\usepackage{graphicx}
\usepackage{caption}

\usepackage{subcaption}
\usepackage{multirow}

\usepackage{enumitem}
\usepackage{svg}
\usepackage{float}
\usepackage[normalem]{ulem}
\usepackage{hyperref}
\usepackage{algpseudocode}

\usepackage{scalerel}
\usepackage{tikz,standalone,pgfplots}
\usepackage{xcolor}
\definecolor{salmon}{RGB}{250, 128, 114}
\usepackage{pifont}
\usepackage{overpic}
\usepackage{balance}
\usepackage[normalem]{ulem} 

\newcommand\redsout{\bgroup\markoverwith
{\textcolor{red}{\rule[0.5ex]{2pt}{0.4pt}}}\ULon}
\newcommand\bluesout{\bgroup\markoverwith
{\textcolor{blue}{\rule[0.5ex]{2pt}{0.4pt}}}\ULon}
\newcommand\purplesout{\bgroup\markoverwith
{\textcolor{purple}{\rule[0.5ex]{2pt}{0.4pt}}}\ULon}
\newcommand\greensout{\bgroup\markoverwith
{\textcolor{green}{\rule[0.5ex]{2pt}{0.4pt}}}\ULon}
\usepackage[normalem]{ulem}

\theoremstyle{definition}
\newtheorem{definition}{Definition}
\AfterEndEnvironment{definition}{\noindent\ignorespaces}

\DeclareMathAlphabet{\pazocal}{OMS}{zplm}{m}{n}

\title{Grasping a Handful: Sequential Multi-Object Dexterous Grasp Generation}
\newacro{dnn}[DNN]{Deep Neural Network}
\newacro{fcn}[FCN]{Fully Convolutional Network}
\newacro{sdf}[SDF]{Signed Distance Function}
\newacro{cnn}[CNN]{Convolutional Neural Network}
\newacro{gnn}[GNN]{Graph Neural Network}
\newacro{dl}[DL]{Deep Learning}
\newacro{ml}[ML]{Machine Learning}
\newacro{mc}[MC]{Monte Carlo}
\newacro{mlp}[MLP]{Multi-Layer Perceptron}
\newacro{dof}[DoF]{Degrees of Freedom}
\newacro{vae}[VAE]{Variational Autoencoder}
\newacro{cvae}[CVAE]{Conditional Variational Autoencoder}
\newacro{methodname}[VCGS]{Variational Constrained Grasp Sampler}
\newacro{fps}[FPS]{Farthest Point Sampling}
\newacro{tai}[TaI]{Target as Input}
\newacro{pca}[PCA]{Principal Component Analysis}
\newacro{pc}[PC]{Principal Component}
\newacro{auc}[AUC]{Area Under the Curve}
\newacro{elbo}[ELBO]{Evidence Lower Bound}
\newacro{bps}[BPS]{Basis Point Set}
\newacro{mala}[MALA]{Metropolis-Adjusted Langevin Algorithm}
\newacro{os}[OS]{Opposition Space}
\newacroplural{os}[OSes]{Opposition Spaces}
\newacro{mcp}[MCP]{Metacarpophalangeal}
\newacro{gan}[GAN]{Generative Adversarial Network}

\definecolor{drakgreen}{rgb}{0.0, 0.5, 0.0}  

\newcommand{\equationref}[1]{\hyperref[#1]{Eq.~\ref*{#1}}}
\newcommand{\figref}[1]{\hyperref[#1]{Fig.~\ref*{#1}}}
\newcommand{\tabref}[1]{\hyperref[#1]{Table~\ref*{#1}}}
\newcommand{\secref}[1]{\hyperref[#1]{Section~\ref*{#1}}}
\newcommand{\algoref}[1]{\hyperref[#1]{Algorithm~\ref*{#1}}}
\newcommand{\lineref}[1]{\hyperref[#1]{Line~\ref*{#1}}}

\newcommand{\defref}[1]{\hyperref[#1]{Definition~\ref*{#1}}}

\newcommand{\matr}[1]{\mathbf{#1}}

\newcommand{\argmin}{\operatornamewithlimits{argmin}}

\newcommand{\etal}[1]{#1 et al.}
\newcommand{\set}[1]{\mathcal{#1}}

\def\methodname{SeqGrasp}
\def\diffusionname{SeqDiffuser}
\def\methodshort{SeqG}
\def\multigraspshort{MulG}
\def\diffusionshort{SeqD}

\def\datasetname{SeqDataset}
\def\graspemshort{G'Em}
\def\graspem{Grasp'Em}
\def\datasetshort{SData}
\def\franka{Franka Emika Panda}

\def\multigrasp{MultiGrasp}

\def\isaacgym{Isaac Gym}

\def\ie{, \textit{i.e.}, }
\def\eg{\textit{e.g.}, }

\author{Haofei Lu$^{\dag 1}$, Yifei Dong$^1$, Zehang Weng$^1$, Florian T. Pokorny$^1$, Jens Lundell$^{1,2}$, and Danica Kragic$^1$
\thanks{Manuscript received: March 27, 2025; Revised: July 10, 2025; Accepted: September 8, 2025. }
\thanks{This paper was recommended for publication by Editor Júlia Borràs Sol upon evaluation of the Associate Editor and Reviewers’ comments.}
\thanks{This work was supported by the Swedish Research Council, the Knut and Alice Wallenberg Foundation, the European Research Council (ERC-BIRD-884807). The authors also would like to express their sincere gratitude to Li Chen for valuable assistance with the experimental hardware.}
\thanks{$^\dag$ Corresponding author.}
\thanks{$^1$ Robotics, Perception, and Learning (RPL) at KTH, Stockholm, Sweden. 
        {\tt\small \{haofeil,yifeid,zehang,fpokorny,jelundel,dani\} \\ @kth.se}}
\thanks{$^2$ Robotics and Autonomous Systems at University of Turku, Turku, Finland.
        {\tt\small jens.lundell@utu.fi}}
        
\thanks{Digital Object Identifier (DOI): see top of this page.}

}
\begin{document}

\maketitle
\begin{abstract}

We introduce the sequential multi-object robotic grasp sampling algorithm \methodname{} that can robustly synthesize stable grasps on diverse objects using the robotic hand's partial \ac{dof}. We use \methodname{} to construct the large-scale Allegro Hand sequential grasping dataset \datasetname{} and use it for training the diffusion-based sequential grasp generator \diffusionname. We experimentally evaluate \methodname{} and \diffusionname{} against the state-of-the-art non-sequential multi-object grasp generation method \multigrasp{} in simulation and on a real robot. The experimental results demonstrate that \methodname{} and \diffusionname{} reach an 8.71\%-43.33\% higher grasp success rate than \multigrasp. Furthermore, \diffusionname{} is approximately 1000 times faster at generating grasps than \methodname{} and \multigrasp. Project page: \url{https://yulihn.github.io/SeqGrasp/}.

\end{abstract}

\section{Introduction}
\label{sec:Introduction}
Dexterous grasping has been extensively studied in robotics \cite{bohg2013data, newbury2023deep}. The field has evolved through two primary research directions. First, numerous methods have been developed for generating robotic grasps \cite{dexdiffuser, lu2023ugg, jensdgcc, wei2024dro, dexgraspnet, li2022gendexgrasp}, from analytical approaches \cite{dexgraspnet} to data-driven generative models including \acp{gan} \cite{jensdgcc}, diffusion models \cite{lu2023ugg, dexdiffuser}, and \acp{vae} \cite{li2022gendexgrasp, wei2024dro}. Second, a complementary line of research has focused on understanding human grasping behaviors, including intention recognition \cite{song2013predicting}, real-time tracking of hand-object interactions \cite{romero2010hands}, grasp taxonomy classification \cite{feix2015grasp}, and learning task-oriented grasping strategies from human demonstrations \cite{kokic2020learning}.

\begin{figure}
    \centering
    \includegraphics[width=0.45\textwidth]{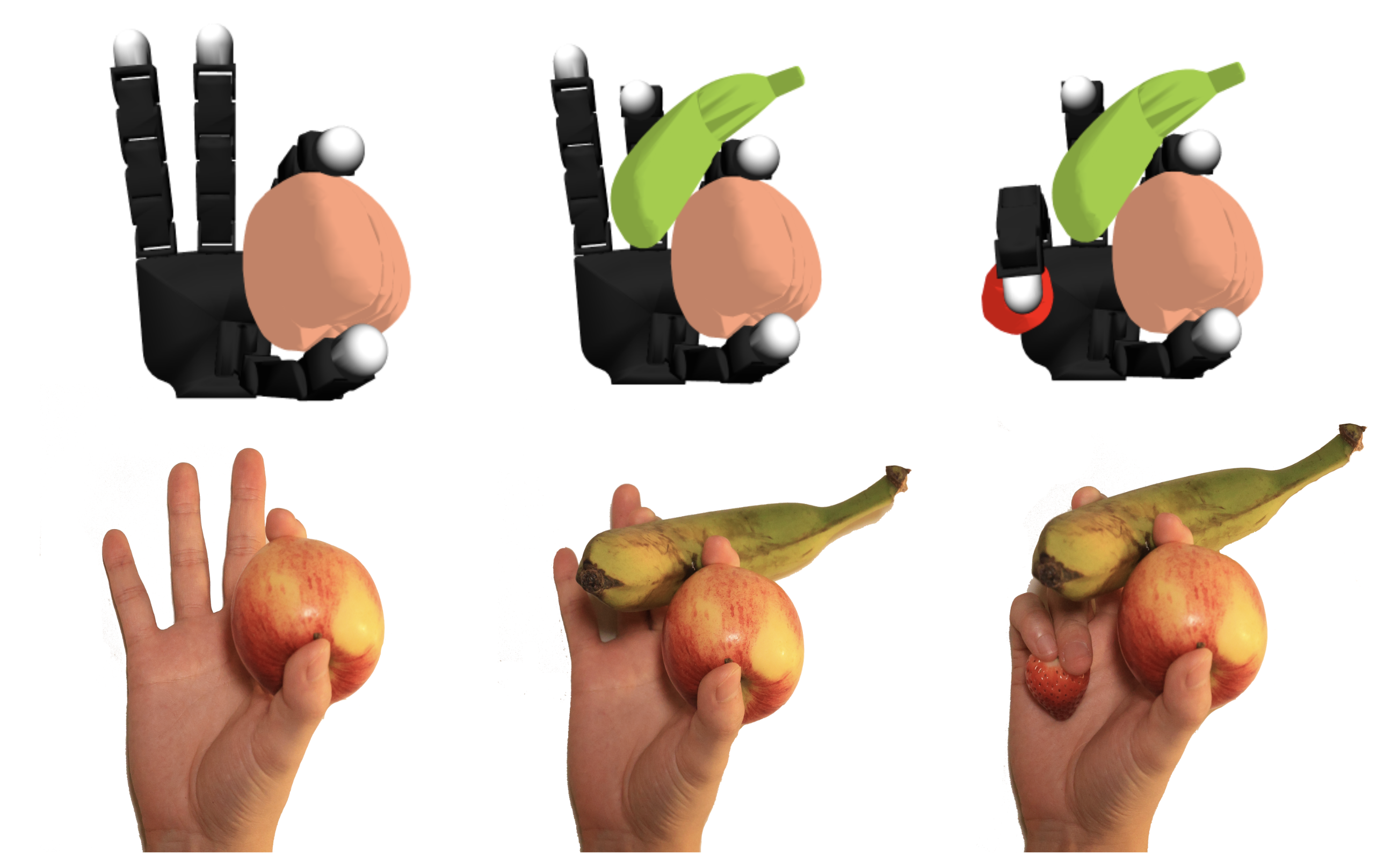}
    \caption{\textbf{Sequential multi-object grasping.}}
    \label{fig:pullimage}
\end{figure}

The primary strategy in dexterous grasp generation has been to bring the robotic hand close to the object and then simultaneously envelop it with all fingers. While this strategy often results in efficient and successful grasp generation, it simplifies dexterous grasping to resemble parallel-jaw grasping, thereby underutilizing the many \ac{dof} of multi-fingered robotic hands~\cite{dexgraspnet}. In contrast, grasping multiple objects with a robotic hand, particularly in a sequential manner that mirrors human-like dexterity, as shown in \figref{fig:pullimage}, is still an unsolved problem.

In this work, we introduce \methodname{}, a novel hand-agnostic algorithm for generating sequential multi-object grasps. Our approach utilizes an optimization-based method to sequentially determine single-object grasp poses using a subset of the hand’s \ac{dof}. As the grasp sequence progresses, the \ac{dof} engaged in previous grasps are frozen, leaving only the remaining \ac{dof} available for subsequent object grasps. To only engage a subset of the hand's \ac{dof} for each grasp, we propose an \ac{os} selection strategy that enables stable grasping using only a pair of links. Using \methodname{}, we construct the large-scale dataset \datasetname{} containing 870K penetration-free Allegro hand grasps across 509 objects, with up to four sequentially grasped objects. Finally, we train the conditional sequential grasping diffusion model \diffusionname{} on \datasetname{} to enable sequential grasping on novel objects.

We experimentally evaluate \methodname{}, \diffusionname{}, and the state-of-the-art simultaneous multi-object grasping method \multigrasp~\cite{multigrasp} in simulation and on physical hardware. The simulation results revealed that \methodname{} and \diffusionname{} perform on par with \multigrasp{} for picking one or two objects while outperforming it when picking three to four objects. Moreover, \diffusionname{} demonstrates superior efficiency, generating 256 grasps within one second compared to approximately 1000 seconds for \methodname{} and \multigrasp. For the real-world experiments, we replicated the grasp sequences proposed by the methods on a real Allegro Hand attached to a Franka Panda Robot. These results align with the simulation findings, demonstrating that \methodname{} reaches a 43.33\% higher grasp success rate than \multigrasp. 

\begin{table*}[h]
    \centering
    \begin{tabular}{c|cccccccc}
        \textbf{Dataset} & Hand & Data Collection & Objects & Obj. Set/Seq. & Grasps & Strategy & Max. Obj. Grasped & Open-source \\[0.5ex] 
        \hline  \\[-6pt]
        
        MOG~\cite{sun2022multi} & Human\&Barrett & Sim./Real  & 11 & \textbackslash{} & 28K & Simu./Seq. & \textbackslash{} & \ding{55} \\
        \graspem{}~\cite{multigrasp} & Shadow & Sim.  & 8 & 36 & 90K & Simultaneous & 2 & \ding{51} \\
        \datasetname{}(Ours) & Allegro & Sim. & 509 & 2400 & 870K & Sequential & 4 & \ding{51} \\
    \end{tabular}

    \caption{\textbf{Multi-object grasping datasets.}}
    \label{tab:datasetcomparison}
\end{table*}
Our contributions can be summarized as follows:
\begin{itemize}
    \item \methodname{}, a novel algorithm for sequential multi-object grasp generation.
    \item  \datasetname{}, a large-scale dataset for sequential multi-object dexterous grasping.
    \item \diffusionname{}, an efficient conditional sequential grasp diffusion model.
    \item Extensive simulation and real-world experiments demonstrating the feasibility and effectiveness of  \methodname{}.
\end{itemize}

\section{Related Work}
\label{sec:related_work}

The work presented here spans dexterous grasping, multi-object grasping, and datasets for dexterous grasping and we review these areas below.

\subsection{Dexterous Grasping}
\textbf{Analytical methods}. Early research in dexterous grasping generated stable grasps by optimizing a grasp quality metric such as the force-closure metric \cite{graspit, old3}. Although these methods are theoretically sound, they are computationally demanding because (i) the many \ac{dof} dexterous hands cause high-dimensional search spaces~\cite{eigengrasp}, and (ii) the quality metrics are expensive to compute \cite{dfc}. Consequently, some methods have focused on reducing the search space by imposing constraints on the hand~\cite{old1,old2} or restricting it to joint configuration subspaces~\cite{eigengrasp}. Another line of work has proposed a computationally cheap differentiable force-closure estimator \cite{dfc, dexgraspnet}, which has the advantage of being hand-agnostic. This work extends the differentiable force-closure measure from \cite{dexgraspnet} to sequential multi-object grasping.

\textbf{Data-driven methods}. Recent advancements in machine learning have significantly improved dexterous grasp generation~\cite{dexdiffuser,mayer2022ffhnet,lu2023ugg,wei2024dro}. Nowadays, deep generative models can generate thousands of dexterous grasps on previously unseen and partially observed objects within seconds~\cite{dexdiffuser,mayer2022ffhnet,lu2023ugg,wei2024dro}, something analytical methods cannot. Still, only a few data-driven dexterous grasping methods have been developed for multi-object grasping \cite{he2025sequential,multigrasp}. In this work, we train a new diffusion-based sequential multi-object grasp sampler \diffusionname{} inspired from \cite{dexdiffuser} on our own optimization-generated sequential multi-object dataset \datasetname.

\subsection{Multi-Object Grasping}

Multi-object grasping presents unique challenges due to the complex multi-object interactions and the high-dimensional configuration space spanned by the hand and the objects. Some prior parallel-jaw multi-object grasping methods~\cite{agboh2022multi,yonemaru2025learning} explored multi-object push grasps where scattered objects are first pushed together to facilitate multi-object grasping. However, these methods are limited by their reliance on simple shape primitives and parallel-jaw grippers. In comparison, our work can handle objects of diverse shapes and sizes.

A few works have addressed dexterous multi-object grasping \cite{multigrasp,he2025sequential,yao2023exploiting} where \cite{multigrasp} targets simultaneous multi-object grasping while \cite{he2025sequential,yao2023exploiting} targets sequential multi-object grasping.~\etal{Li} \cite{multigrasp} proposed \multigrasp{} a two-stage simultaneous multi-object dexterous grasping framework where a generative grasp sampler proposed poses to simultaneously pick many objects, followed by a learned policy for executing the pick. Simultaneously grasping multiple objects is an efficient strategy because it avoids repeated hand repositioning, which reduces total execution time, but the main limitation is that objects must be spatially close and of similar size and shape. In comparison, our method can handle scattered objects of different shapes and sizes by sequentially picking one at a time. The other works that do sequential multi-object grasping \cite{he2025sequential,yao2023exploiting} restrict the grasping to a maximum of two objects \cite{he2025sequential} or to primitive object shapes such as cylinders or spheres \cite{yao2023exploiting}. In comparison, our method can handle up to four objects of complex shapes and sizes. We also use our method to collect the largest sequential multi-object grasping dataset to date.

\subsection{Datasets for Dexterous Grasping}

Large-scale dexterous grasp datasets~\cite{jensdgcc, dexgraspnet, chen2024bodex,sun2022multi, multigrasp,mayer2022ffhnet, graspxl,realdex} have significantly advanced the training of data-driven dexterous grasping methods. However, most of these datasets target single-object grasping \cite{jensdgcc, dexgraspnet, chen2024bodex, mayer2022ffhnet, graspxl,realdex}, with only a few for multi-object grasping~\cite{sun2022multi, multigrasp}. As shown in \tabref{tab:datasetcomparison}, these existing multi-object datasets are small and predominantly focus on simultaneous rather than sequential grasping. Therefore, we collect the new large-scale sequential multi-object grasping dataset \datasetname{}\footnote{The dataset is available at \url{https://yulihn.github.io/SeqGrasp/}.} using our method \methodname. \datasetname{} is, to date, by far the largest multi-object grasping dataset.

\section{Problem Formulation}
\label{sec:problem_formulation}
The problem addressed in this work is sequential multi-object grasping, which we define as follows:
\begin{definition}[Sequential multi-object grasping]
A sequential multi-object grasp is a grasp where one object is grasped at a time using a subset of the dexterous hand's \ac{dof}, while previously grasped objects, if any, remain fixed to the hand.
\end{definition}

\noindent To contrast, \textit{simultaneous multi-object grasping} addresses how to grasp multiple objects \textit{simultaneously}, typically utilizing all the \ac{dof} of the hand \cite{multigrasp}.

We formulate the sequential multi-object grasping problem as generating a sequence of $N$ grasps $\set{G}=(\matr{g}_i)_{i=1}^N$ for picking a sequence of $N$ objects $\set{O}=(\matr{O}_i)_{i=1}^{N}$, where ${N \geq 2}$ and $(\cdot)$ denotes an ordered sequence. Each $\matr{g}_n\in\set{G}$ is restricted to a specific subset $\mathcal{OS}_n$ of the hand's total \ac{dof}. Mathematically, this can be described as
\begin{align}
    \argmin_{\matr{g}_n} E(\matr{g}_{n}, \matr{O}_{n}, \set{G}_{n-1}, & \; \set{O}_{n-1}, \mathcal{OS}_n), \forall n= 1,\dots,N,
    \label{eq:problem_formulation}
\end{align}
where $\set{G}_{n-1}=(\matr{g}_i)_{i=1}^{n-1}$, $\set{O}_{n-1}=(\matr{O}_i)_{i=1}^{n-1}$, $\set{G}_{0}=\varnothing$, and $\set{O}_{0}=\varnothing$. $E$ in \equationref{eq:problem_formulation}  is a differentiable function that quantifies how well grasp $\matr{g}_n$ can pick object $\matr{O}_n$ with the \ac{dof} $\mathcal{OS}_n$ given all previously generated grasps $\set{G}_{n-1}$  and objects $\set{O}_{n-1}$. We assume all previously grasped objects $\set{O}_{n-1}$ remain fixed in the hand when generating a grasp on the next object $\matr{O}_n$.

In this work, we represent $\mathcal{OS}_n$ as an opposition space (\secref{subsec:grasp_selection_strategy}), each object $\matr{O}\in \set{O}$ as a triangular mesh, and each grasp $\matr{g}\in \set{G}$ as a vector $\matr{g}=\left[\matr{p}, \matr{r}, \boldsymbol{\theta}\right]\in\mathbb{R}^{9+K}$, where $\matr{p}\in\mathbb{R}^3$ is the hand's base position, $\matr{r}\in \mathbb{R}^6$ is the hand's base orientation in a 6D continuous reprsentation~\cite{6drot}, and ${\boldsymbol{\theta}\in\mathbb{R}^K}$ is the $K$-dimensional hand joint angles which are 16 for the Allegro Hand. We assume the shape of all objects in $\set{O}$ to be fully known. Next, we will introduce \methodname{} our algorithm for solving \equationref{eq:problem_formulation}.

\section{Sequential Grasp Generation}
\label{sec:sequential_grasp_generation}

\begin{figure}[b]
    \centering
    \includegraphics[width=0.8\linewidth]{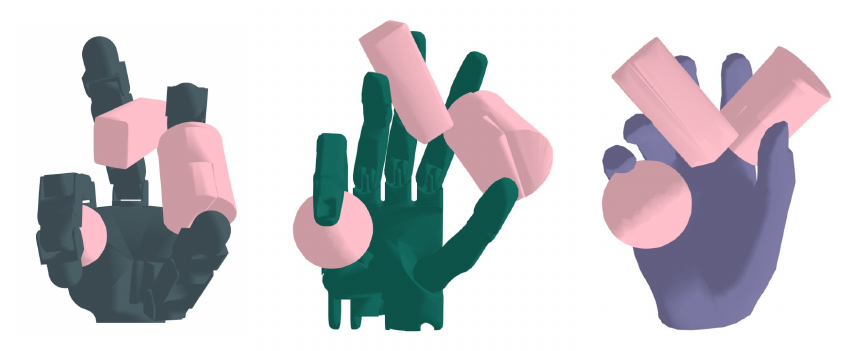}
\caption{\textbf{Multi-object grasping visualizations} for different hands. From left to right: Allegro Hand, Shadowhand, MANO.}
\label{fig:multihand}
\end{figure}

\begin{algorithm}[!ht]
    \caption{\methodname{}}\label{alg:alg}
    \footnotesize
    \SetAlgoLined
    \SetKwInOut{Input}{Input}\SetKwInOut{Output}{Output}
    
    \Input{Object sequence $\set{O}$, \acp{os} $\mathcal{OS}$, $N_{\text{step}}$, and $p_{\text{accept}}$.}

    \Output{The optimized grasp sequence $\set{G^*}_n$.}

    $n=1$;

    \vspace{3pt}
    
    \While{ $\mathcal{OS}\neq \varnothing$ and $n \leq N$}{
    \vspace{2pt}
    
        $\mathcal{OS}_n \sim \mathcal{U}(\mathcal{OS})$;\label{op:sampling}
        
    $\{\matr{x}_j\}_{j=1}^{2} \sim \mathcal{U}(\matr{S}_n)$;\label{op:contactpointsinit}

    \For{$\text{s}=1$ \KwTo $N_{\text{step}}$}{ \label{op:for_loop_start}

        $\Delta = \partial E(\matr{g}_{n}, \matr{O}_{n}, \set{G}_{n-1}, \set{O}_{n-1}, \{\matr{x}_j\}_{j=1}^{2}) / \partial\matr{g}_{n} $;
        
        $\matr{g}_{n} \leftarrow \text{MALA}(\matr{g}_{n}, \matr{J}_n, \Delta)$ ; \label{op:mala}
        
        $\{\matr{x}_j\}_{j=1}^{2} \sim f(\matr{S}_n, p_{\text{accept}})$;\label{op:contactpointsresamp}
    } \label{op:for_loop_end}
    
    $\mathcal{OS} \gets \mathcal{OS} \setminus \mathcal{OS}_n$;\label{op:removing}

    \For{$\mathcal{OS}_j \in \mathcal{OS}$}{\label{op:update_starts}
        $\matr{J}_j \gets \matr{J}_j\odot (\mathbf{1}-\matr{J}_n)$;\label{op:J_update}
            
        \If{$\matr{J}_j = \mathbf{0}$}{
            
            $\mathcal{OS} \gets \mathcal{OS} \setminus \mathcal{OS}_j$;\label{op:occupied}
        }
    }\label{op:update_ends}

    $n \mathrel{+}= 1$;
    }
\end{algorithm}

Here, we present \algoref{alg:alg} for sequential grasp generation. It includes (i) an opposition space selection strategy (\secref{subsec:grasp_selection_strategy}), (ii) an optimization-based grasp synthesis method (\secref{subsec:optimization_grasp_generation}), and (iii) an energy-based cost function (\secref{sec:energy_function}). \figref{fig:multihand} shown an example of running \algoref{alg:alg} to grasp three different objects with three different dexterous hands.

\subsection{Opposition Space Selection Strategy}
\label{subsec:grasp_selection_strategy}

\begin{figure}[b]
    \centering
     \begin{subfigure}[b]{0.28\textwidth}
         \centering
         \includegraphics[width=\textwidth]{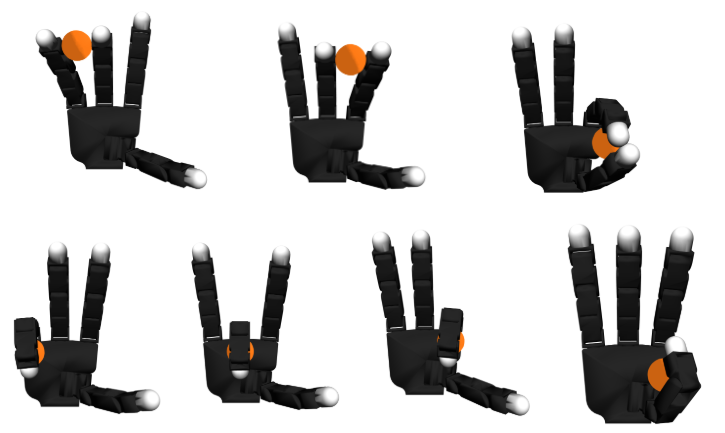}
         \caption{}
         \label{fig:grasptypes}
     \end{subfigure}
     \hfill
    \begin{subfigure}[b]{0.18\textwidth}
             \centering
             \includegraphics[width=\textwidth]{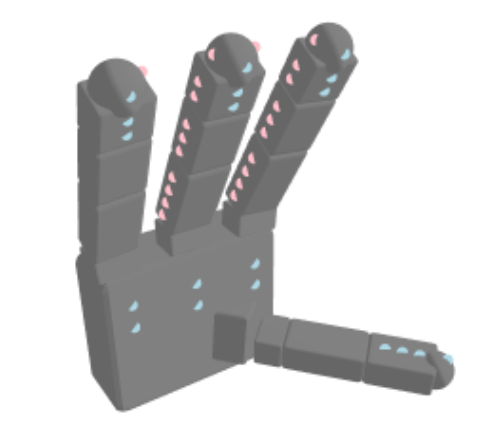}
             \caption{}
             \label{fig:allegro_contacts}
    \end{subfigure}     
\caption{(a) \textbf{Grasps using all seven \ac{os}es}. From left to right, first row: middle-ring, index-middle, and thumb-index, second row: ring-palm, middle-palm, index-palm, and thumb-palm. (b) \textbf{Visualization of contact point candidates} on Allegro Hand surface. \textcolor{cyan}{Cyan} and \textcolor{pink}{pink} points denote palm opposition and side opposition contacts, respectively. }
\end{figure}

The primary objective in sequential multi-object grasping is to maximize the hand's remaining \ac{dof} after each grasp. For this purpose, we propose a grasp planning strategy guided by \acp{os}~\cite{iberall1986opposition, feix2015grasp, smeets2019review, yao2023exploiting}. An \ac{os} is a functional subspace within the hand's kinematic structure formed by pairs of opposing surfaces (such as fingertips, lateral surfaces of fingers, or palm surfaces) along with the joints that control these surfaces \cite{yao2023exploiting}. It represents regions where opposing forces can be applied to create stable grasps. The number of \acp{os} are hand-dependent and vary based on the kinematic structure. \figref{fig:grasptypes} shows the seven different \acp{os} for the Allegro Hand.

Mathematically, each opposition space can be represented as a pair $\mathcal{OS}_i = \{\matr{J}_i, \matr{S}_i\}$, where $\matr{J}_i \in \{0,1\}^{K}$ is a binary vector indicating which joints are involved in controlling the opposition space, and $\matr{S}_i\in \mathbb{R}^{3\times M_i}$ represents the 3D points on the hand where opposing forces can be applied. \figref{fig:allegro_contacts} shows an example of two different $\matr{S}_i$ for the Allegro Hand, where palm and pad oppositions have contact points located on the inner surfaces of fingers and palm and side oppositions have contact points on the fingers’ lateral surfaces.

Let $\mathcal{OS}=\{\mathcal{OS}_i\}_{i=1}^L$ be the set of all \acp{os}. Given this set, \algoref{alg:alg} samples a random \ac{os} from it (\lineref{op:sampling}) and uses it for subsequent grasp generation (\secref{subsec:optimization_grasp_generation}).
Once grasp generation is complete, the sampled \ac{os} can no longer be used and is thus removed from the available \acp{os} (\lineref{op:removing}).  $\matr{J}_i$ of all the remaining \acp{os} are also updated by zeroing out the joints used in $\mathcal{OS}_{n}$ (\lineref{op:J_update}). Subsequently, all \acp{os} with $\matr{J}=\mathbf{0}$, meaning that no more controllable joints exist, are removed (\lineref{op:occupied}). For instance, in the case of the Allegro Hand, if the thumb-index \ac{os} is selected, then both the thumb-palm and index-palm \acp{os} become unavailable due to shared joint constraints.

\begin{figure}[t]
    \centering
\begin{subfigure}[b]{0.15\textwidth}
         \centering
         \includegraphics[width=\textwidth]{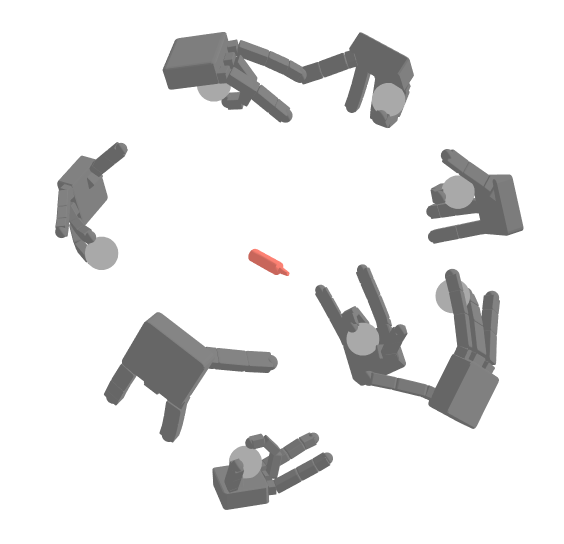}
         \caption{}
         \label{fig:initialization}
     \end{subfigure}
     \hfill
     \begin{subfigure}[b]{0.30\textwidth}
         \centering
         \includegraphics[width=\textwidth]{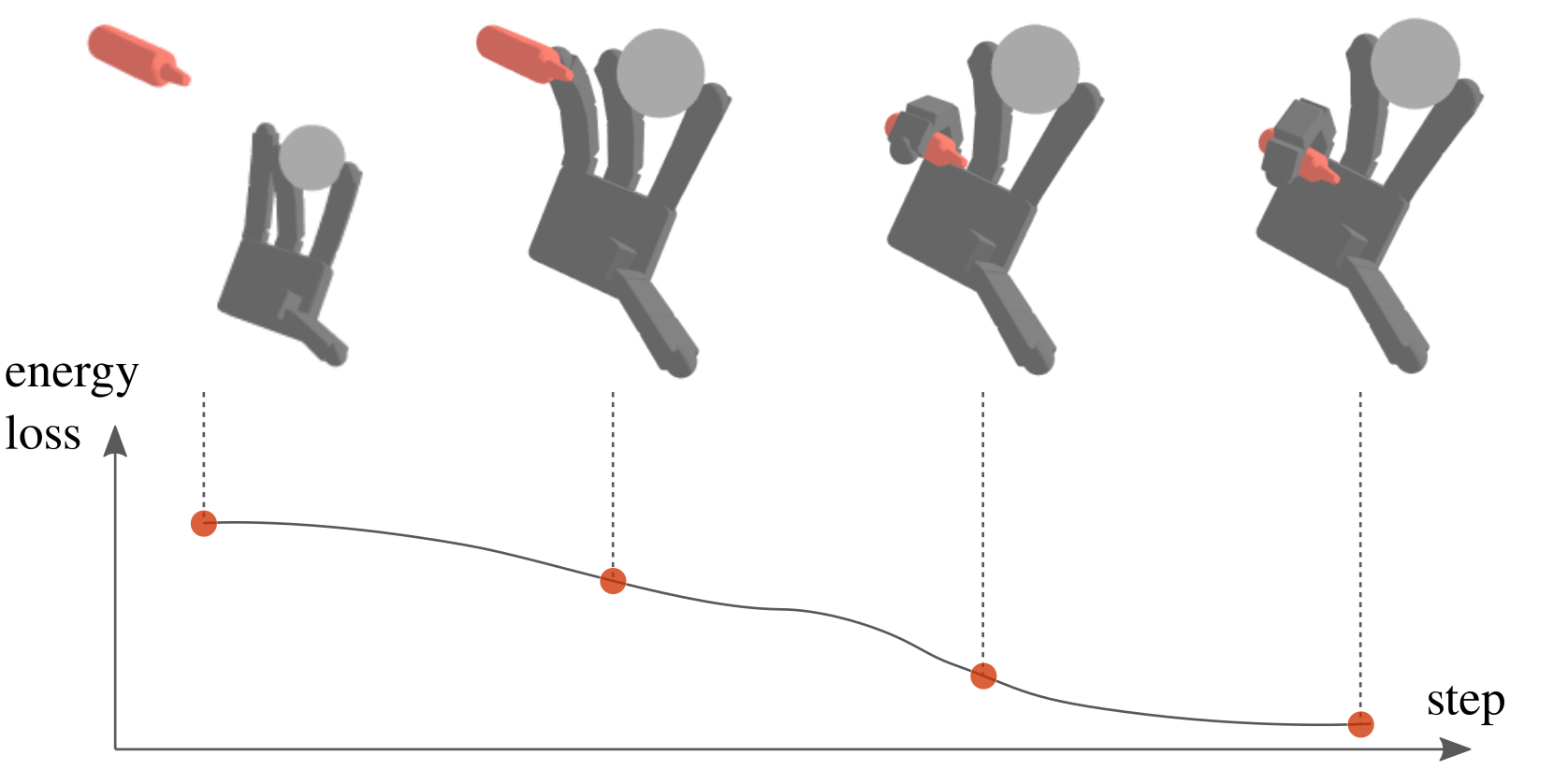}
         \caption{}
         \label{fig:optimization}
     \end{subfigure}
\caption{(a) \textbf{Initialization.} The initial grasp configurations are randomly sampled on the expanded convex hull of the object (\textcolor{salmon}{bottle}) while the previously grasped object (\textcolor{gray}{ball}) remains in the hand. (b) \textbf{Optimization.} During optimization, the grasp is incrementally refined, ultimately securing the target object using the ring-palm \ac{os}.}
\end{figure}
\subsection{Optimization-based Grasp Generation}
\label{subsec:optimization_grasp_generation}

The next step in the algorithm (Lines \ref{op:for_loop_start}-\ref{op:for_loop_end}) is to generate a stable, physically plausible, and collision-free grasp that respects 
the sampled $\mathcal{OS}_n$. To achieve this, we formulate $E$ in \equationref{eq:problem_formulation} as an energy function (\secref{sec:energy_function}) and numerically optimize it using the \ac{mala} \cite{Roberts1996ExponentialCO} (\lineref{op:mala}).

In robotic grasping, \ac{mala} has been used to optimize single object grasps $\matr{g}_n$ by iteratively refining ~$\matr{p}_n$, $\matr{r}_n$ and $\boldsymbol{\theta}_n$ according to Langevin dynamics \cite{dfc,dexgraspnet}.  However, we must adapt \ac{mala} to sequential multi-object grasping. To this end, we propose a new grasp as $\hat{\matr{g}}_{n} \leftarrow \matr{g}_{n} - \gamma \left[\mathbf{1},\matr{J}_n\right]\odot\Delta$, where $\gamma$ is the step size, $\mathbf{1} \in \mathbb{R}^9$ is a padding vector to align the length of $\matr{J}$ with $\matr{g}$, $\Delta = \partial E / \partial\matr{g}_{n}$ is the energy gradient, and $\odot$ is the element-wise (Hadamard) product. $\hat{\matr{g}}_{n}$ is accepted if $\alpha\geq u$, where ${u\sim \mathcal{U}([0,1])}$ and 
\begin{equation}
\alpha=\frac{E(\hat{\matr{g}}_{n}, \matr{O}_{n}, \set{G}_{n-1}, \set{O}_{n-1}, \{\matr{x}_j\}_{j=1}^{2})}{E(\matr{g}_{n}, \matr{O}_{n}, \set{G}_{n-1}, \set{O}_{n-1}, \{\matr{x}_j\}_{j=1}^{2})}.
\label{eq:alpha}
\end{equation}

The above procedure is repeated for a fixed number of steps where, at each step, $\{\matr{x}_j\}_{j=1}^{2}$ is re-sampled with probability $p_{\text{accept}}$ (\lineref{op:contactpointsresamp}). This resampling process helps accelerate convergence and escape from local minimas~\cite{dfc,dexgraspnet}.

We initialize $\matr{g}_n$ at a randomly sampled position on the expanded convex hull of the target object $\matr{O}_n$ as exemplified in \figref{fig:initialization},. If $n=1$, then $\boldsymbol{\theta}_1$ is set to a natural open-hand and collision-free posture, as demonstrated in \figref{fig:allegro_contacts}, all fingers are fully extended to create sufficient space for grasping objects with any \ac{os}. While for $n \geq 2$, $\boldsymbol{\theta}_n=\boldsymbol{\theta}_{n-1}$. A visual example of the optimization process when grasping a second object is shown in \figref{fig:optimization}.

\subsection{Energy Function}
\label{sec:energy_function}
Numerically optimizing the energy function in \equationref{eq:problem_formulation} should result in stable, collision-free, joint-respecting, and \ac{os}-respecting grasps. We design the following energy function to capture all of these behaviors   
\begin{equation}
    E = \matr{w}^T 
    \begin{bmatrix}
        E_{\text{fc}} & E_{\text{dis}} & E_{\text{hop}} & E_{\text{hsp}} & E_{\text{joint}} & E_{\text{oop}}
    \end{bmatrix}^T,
\label{eq: energy-function}
\end{equation}
where $\matr{w}\in\mathbb{R}^6$ is a weight vector controlling the relative importance of the force-closure $E_{\text{fc}}$, contact distance $E_{\text{dis}}$, hand-object penetration $E_{\text{hop}}$, hand self-penetration $E_{\text{hsp}}$, joint limits $E_{\text{joint}}$, and object-object penetration $E_{\text{oop}}$ energy terms. 

The force-closure term ($E_{\text{fc}}$) encourages the grasp to be in force-closure equilibrium~\cite{rimon1996force}. The definition for ($E_{\text{fc}}$) originates from~\cite{dexgraspnet}, where zero friction and uniform contact force magnitudes are assumed because it significantly speeds up the computation. We follow the same definition and write it as
\begin{align}
\label{eq:Ec}
    E_{\text{fc}}(\{\matr{x}_j\}_{j=1}^{2}) = \lVert \matr{G}\matr{c} \rVert^2,
\end{align}
where ${\matr{c} = [\matr{c}_1^T,\matr{c}_2^T]^T \in \mathbb{R}^{6 \times 1}}$ represents the concatenated contact normals at each contact point~$\{\matr{x}_j\}_{j=1}^2$. $\matr{G}$ is defined as:
\begin{align}
\label{eq:Ec_explain}
    \matr{G} = \begin{bmatrix} 
        \matr{I} & \matr{I} \\ 
        [\matr{x}_1]_\times &  [\matr{x}_2]_\times \\ 
    \end{bmatrix},
\end{align}
\noindent where $\matr{I}$ represents the identity matrix, and $[\matr{x}_j]_\times$ (${1 \leq j \leq 2}$) denotes the skew-symmetric matrix formed from the contact point $\matr{x}_j$.

The contact distance and penetration terms ($E_{\text{dis}}$ \& $E_{\text{hop}})$ encourage the hand-object contacts to occur close to the object surface without penetrating it. The contact distance is mathematically defined as
\begin{align}
E_{\text{dis}}(\{\matr{x}_j\}_{j=1}^{2}, \matr{O}_n) = \sum_{j=1}^{2} d(\matr{x}_j,\matr{O}_n),
\end{align}
where ${d(\matr{x}_j,\matr{O}_n)=\min_{\matr{v} \in \matr{O}_n} {\lVert \matr{x}_j -\matr{v} \rVert}_2}$ is the shortest point-mesh distance. Similarly, the hand-object penetration term is defined as:
\begin{align}
E_{\text{hop}}(\matr{g}_n, \matr{O}_n) &= \sum_{\matr{v} \in \mathcal{V}_{\text{hop}}(\matr{H}_\matr{g}, \matr{O}_n)} d(\matr{v},\matr{O}_n),
\end{align}
where ${d(\matr{v},\matr{O}_n)=\min_{\matr{v}_1 \in \matr{O}_n} {\lVert \matr{v} -\matr{v}_1 \rVert}_2}$ and $\mathcal{V}_{\text{hop}}(\matr{H}_{\matr{g}}, \matr{O}_n)$ is the set of points on the hand surface pointcloud ${\matr{H}_{\matr{g}} \in \mathbb{R}^{3 \times M_h}}$ that penetrate the object $\matr{O}_n$.

The self-collision and joint limit terms ($E_{\text{hsp}}$ \& $E_{\text{joint}}$) encourage physical feasibility. We define these as 
\begin{align}
E_{\text{hsp}}(\matr{g}_n) &= \sum_{\matr{v}_1,\matr{v}_2 \in \mathcal{V}_{\text{hsp}}(\matr{H}_{\matr{g}}), \matr{v}_1 \neq \matr{v}_2} \max({\lVert \matr{v}_1 - \matr{v}_2 \rVert}_2, 0), \\
E_{\text{joint}}(\matr{g}_n) &= {\|(\boldsymbol{\theta}-\boldsymbol{\theta}^{\text{upper}})^+\|}_1 + {\|(\boldsymbol{\theta}^{\text{lower}}-\boldsymbol{\theta})^+\|}_1,
\label{eq:energy-feasibility}
\end{align}
where $\mathcal{V}_{\text{hsp}}(\matr{H}_{\matr{g}})$ denotes all surface points of the hand that are self-penetrating, $(\cdot)^+$ denotes the element-wise operation $\max(\cdot, 0)$, 
and $\boldsymbol{\theta}^{\text{upper}}$ and $\boldsymbol{\theta}^{\text{lower}}$ denote the upper and lower limits of all joints.

Finally, the term ($E_{\text{oop}}$) minimizes object-object penetration. It is defined as
\begin{equation}
E_{\text{oop}}(\{\matr{O}_i\}_{i=1}^{n}) = \sum_{i=1}^{n-1} \sum_{\matr{v} \in \mathcal{V}_{\text{oop}}(\matr{O}_i, \matr{O}_{n})} d(\matr{v},\matr{O}_n),
\end{equation}
where ${d(\matr{v},\matr{O}_n)=\min_{\matr{v}_1 \in \matr{O}_n} {\lVert \matr{v} -\matr{v}_1 \rVert}_2}$, and  $\mathcal{V}_{\text{oop}}(\matr{O}_i, \matr{O}_{n})$ are the inter-penetrating surface points between the previously grasped object $\matr{O}_i$ and the current object $\matr{O}_n$.

\section{Dataset Generation}
\label{datasetcuration}

We use \methodname{} to generate our large-scale dataset \datasetname{} \footnote{The dataset is available at \url{https://yulihn.github.io/SeqGrasp/}.} containing 4.9 million sequential Allegro Hand grasps on over 509 objects from DexGraspNet~\cite{dexgraspnet}. Each object is resized to fit within the Allegro Hand by scaling its axis-aligned bounding box to be between 0.06 and 0.10 meters. Then, from the 509 resized objects, we generate 600 unique object sets containing four objects each. The objects in these sets are randomly permuted four times, resulting in 2,400 unique object sequences.

We run \algoref{alg:alg} on the 2,400 unique object sequences with $\matr{w} = [50, 50, w_{\text{hop}}, 5, 1, 5]^T$, where $w_{\text{hop}}$, that penalizes hand-object penetration, grows linearly from $5$ to $5e^2$. The optimization runs for 6,000 iterations per grasp. We validate the generated grasp sequences in the physics simulator \isaacgym{} following the setup in~\cite{dexgraspnet}. This setup initializes all object densities to 500 $\text{kg/m}^3$, the friction coefficient to 2.0, the hand to the generated grasp configuration, and objects as free-floating. Then, the hand closes until contact is established with the object, which happens when the distance between the hand and the object is less than 2 mm. Finally, a grasp stability test is evaluated by linearly applying an acceleration of 9.8 $\text{m}/\text{s}^2$ in all six orthogonal directions for 100 consecutive simulation steps. A grasp sequence is successful if all grasped objects remain in contact with the hand after the grasp stability test and the maximum penetration depth is less than 1 cm. Otherwise, the grasp sequence is a failure. This procedure ultimately produced 4.9 million grasps, of which 870K (17.82\%) were successful. Only successful grasps were retained in \datasetname{}. The entire data generation process requires approximately 2,500 GPU hours on an NVIDIA A100.

The statistics of \datasetname{} are presented in \tabref{tab:datasetstats}. The results demonstrate that grasps using palm opposition achieved significantly higher success rates, suggesting that these \acp{os} are important in sequential multi-object grasping. Notably, grasps using thumb-index and thumb-palm \acp{os} display lower success rates than other \acp{os}, which deviates from previous research findings that underscore the thumb's dominant role in human hand manipulation tasks~\cite{thumb}. We hypothesize that this discrepancy may stem from the biomechanical differences between the Allegro and the human hand. Finally, \datasetname{}’s grasp consumption, which indicates no more available \acp{os} ($\mathcal{OS}=\emptyset$), aligns with its objective of supporting multi-object grasping, with a significant portion of cases (96.12\%) showing potential for sequential grasping of three to four objects, indicating efficient utilization of the hand's \ac{dof}.

\begin{table}[t]
    \centering
    \begin{adjustbox}{max width=0.9\linewidth}
    \begin{tabular}{c|ccc}
        \textbf{\ac{os}es}  & \textbf{Success} & \textbf{Total} & \textbf{Success Rate (\%)} \\
       \hline \\[-6pt]
       Middle-Ring   & 100.83 & 704.97& 14.30  \\
       Index-Middle & 97.43 & 700.90   &  13.90 \\
       Thumb-Index & 97.29 & 1691.41   &  5.75 \\
       Ring-Palm &  147.33 & 323.31  &  45.57 \\
       Middle-Palm & 152.77 & 404.84  &  37.74 \\
       Index-Palm &  194.13 & 546.60  &  35.52 \\
       Thumb-Palm &  85.86 & 485.80   &  17.67\\
       \hline \hline \\[-6pt] 
       \textbf{Num. Obj. Grasped}  & \textbf{Consumed} & \textbf{Total} & \textbf{Consumed Rate (\%)} \\
       One   &0.00     & 444.72  &  0.00  \\
       Two   &9.75  & 251.52  &  3.88  \\
       Three &69.20 & 139.31  &   49.68  \\
       Four  &40.09 & 40.09   & 100.00 \\
    \end{tabular}
    \end{adjustbox}

    \caption{\textbf{Statistics of \datasetname{}.} All numbers of grasps are shown in thousand. }
    \label{tab:datasetstats}
\end{table}

\section{Conditional Sequential Grasp Diffuser}\label{sec:Seqdiffuser}
Finally, we introduce \diffusionname{}, a diffusion-based sequential grasp generation method trained on \datasetname{}. The architecture of \diffusionname{} is similar to the sampler from \cite{dexdiffuser}, with the distinction that \diffusionname{} is also conditioned on the \ac{os} to enable sequential multi-object grasping.

For generating the grasp~$\matr{g}_n$ on the $n$-th object~$\matr{O}_n$, we first sample $\mathcal{OS}_n \sim \mathcal{OS}$. The selected \ac{os} is encoded as a one-hot feature vector~$\matr{f}_{n} \in \{\matr{f}^i\}_{i=1}^L$. This feature vector is concatenated with the grasp $\matr{g}_n$ forming the augmented grasp representation $
    \matr{\tilde{g}}_n = [\matr{f}_{n}, \matr{p}_n, \matr{r}_n, \boldsymbol{\theta}_n]$. 
Of all the elements in $\matr{\tilde{g}}_n$, we only want to diffuse $\matr{p}_n$, $\matr{r}_n$, and the subset of $\boldsymbol{\theta}$ that corresponds to $\set{OS}_n$\ie{}$\matr{J}_n$. Therefore, we create the binary diffusion selection vector $\matr{k}_n = \left[\mathbf{0}, \mathbf{1}, \matr{J}_n \right]$, where $\mathbf{0}\in\mathbb{R}^L$ and $\mathbf{1}\in\mathbb{R}^9$. Note that in the following, the subscript $t$ of $\matr{g}_t$ denotes the timestep in the diffusion process, while the sequential grasp step $n$ is omitted from $\matr{\tilde{g}}_n$ and $\matr{k}_n$ for clarity.

The forward process for adding noise to a successful grasp $\matr{\tilde{g}}_0$ over $T$ timesteps is
\begin{align}
\label{eq:forward_diff}
    q(\matr{\tilde{g}}_{1:T}|\matr{\tilde{g}}_0) &= \prod_{t=1}^{T} q(\matr{\tilde{g}}{_{t}}|\matr{\tilde{g}}_{t-1}), \\
    q(\matr{\tilde{g}}{_{t}}|\matr{\tilde{g}}{_{t-1}})&=\mathcal{N}(\matr{\tilde{g}}{_{t}};\matr{k} \odot \sqrt{1-\beta_{t}}\matr{\tilde{g}}{_{t-1}}, \beta_{t} \matr{I}),
\end{align}
where $\beta_{t}$ is the scheduled noise variance at time step $t$. To reconstruct the original $\matr{\tilde{g}}_0$ from $\matr{\tilde{g}}_T$, \diffusionname{} learns the reverse diffusion process by estimating the Gaussian noise at each step $t$ and progressively removing it:
\begin{align}
\label{eq:inv_diff}
    p_{\boldsymbol{\psi}}(\matr{\tilde{g}}{_{0:T}}|{\matr{h}_{\matr{O}}}) &= p(\matr{\tilde{g}}{_{T}})\prod_{t=1}^{T} p_{\boldsymbol{\psi}}(\matr{\tilde{g}}{_{t-1}}|\matr{\tilde{g}}{_{t}}, {\matr{h}_{\matr{O}}}), \\
    p_{\boldsymbol{\psi}}(\matr{\hat{g}}{_{t-1}}|\matr{\tilde{g}}{_{t}}, \matr{h}_{\matr{O}}) &= 
    \mathcal{N}(\matr{\tilde{g}{_{t-1}}}; \matr{k} \odot \hat{\mu}_{\boldsymbol{\psi}}(\matr{\tilde{g}}_t,\matr{h}_{\matr{O}},t),  \notag \\ &\quad\quad\quad\quad\quad \hat{\Sigma}_{{\boldsymbol{\psi}}}(\matr{\tilde{g}}_t,\matr{h}_{\matr{O}},t)),
\end{align}
where $\hat{\mu}_{\boldsymbol{\psi}}(\matr{\tilde{g}}_t,\matr{h}_{\matr{O}},t)$ and $\hat{\Sigma}_{{\boldsymbol{\psi}}}(\matr{\tilde{g}}_t,\matr{h}_{\matr{O}},t)$ are the learnable mean and variance of a Gaussian distribution and $\matr{h}_{\matr{O}}$ is the \ac{bps}~\cite{prokudin2019efficient} encoded feature of the target object~$\matr{O}$. In the forward and reverse processes, noise is only added or removed at positions where $\matr{k}$ is nonzero. The loss function is thereby formulated as:
\begin{align}
\label{eq:noise_pred}
    \mathcal{L}_{\epsilon} = ||\matr{k} \odot \hat{ \matr{\epsilon}}{_{t}} - \matr{k} \odot \matr{\epsilon}{_{t}}||^2,
\end{align}
where $\hat{\matr{\epsilon}}{_{t}}=\epsilon_{\psi}(\matr{\tilde{g}}_t,\matr{h}_{\matr{O}},t)$ and $\matr{\epsilon}_{t}$ are the estimated noise and ground-truth noise, respectively.

\section{Experiments}

\begin{figure}[!t]
    \centering

\begin{tikzpicture}
        \node[anchor=south west, inner sep=0] (image) at (0,0) {\includegraphics[width=0.5\textwidth]{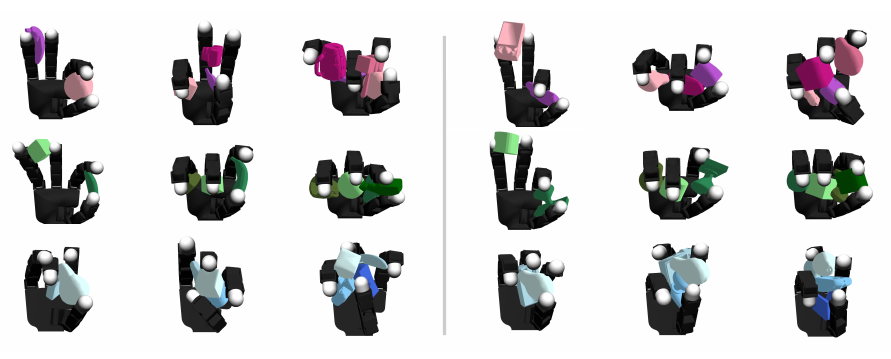}};
        \begin{scope}[x={(image.south east)}, y={(image.north west)}] 
            \node[black] at  (0.25,1.0) {\datasetname{}};
            \node[black] at (0.75,1.0) {\graspem{}};
        \end{scope}
    \end{tikzpicture}
\caption{\textbf{Qualitative results.} For \textcolor[rgb]{0.7,0,0}{\methodname{}} (first row) and \textcolor[rgb]{0,0.5,0}{\diffusionname{}} (second row), the grasp sequences are visually indicated by a color gradient, transitioning from lighter to darker shades. In contrast, for \textcolor[rgb]{0,0,00.9}{\multigrasp{}} (third row), the color gradient is only used to differentiate the objects. For \methodname{} and \diffusionname{}, we only show consumed grasps, that is, when $\mathcal{OS}=\emptyset$.}
\label{fig:sim_visual_result}
\end{figure}

We experimentally evaluate \methodname{} and \diffusionname{} in both simulation and the real world. The specific questions we aim to address with the experiments are:
\begin{enumerate}
    \item How well can \methodname{} and \diffusionname{} generate successful and diverse grasps?
    \item What is the difference between simultaneously and sequentially grasping multiple objects?
    \item Are the generated grasps executable on real hardware?
\end{enumerate}

We compare our method to the optimization-based sampler \multigrasp{} from~\cite{multigrasp}, which generates simultaneous grasps on clustered objects. As such, for \multigrasp{}, we must first sample clustered object configurations and then generate multi-object grasps directly on the object cluster. In contrast, \methodname{} and \diffusionname{} do not require objects to be spatially close as they generate grasps sequentially based on previously successful ones. While this comparison is not entirely fair, we believe that comparing these two strategies still offers valuable insights. 

For training \diffusionname{}, we split \datasetname{} into an 80\% training set and a 20\% test set, ensuring that no training objects were used in the experimental evaluation. The object point clouds are obtained by sampling 2048 points on the object mesh surfaces using \ac{fps}.

\subsection{Simulation Experiments}
\label{simulationexp}

We evaluated all generated grasps in the simulation experiments in \isaacgym{}\cite{makoviychuk2021isaac}. We used two object sets: (1) all eight objects from Grasp'Em~\cite{multigrasp} and (2) a random selection of eight validation objects from \datasetname{}. We randomly generated ten four-object sequences for each object set, and, per object, we generated 256 grasps, resulting in 10,240 grasps per method. 

We used the following metrics to assess the quality of the generated grasps:
\begin{enumerate}
    \item \textbf{Success rate (SR) in percent}:  The same success criteria as described in \secref{datasetcuration}.
    
    \item \textbf{Maximum penetration depth (Pene.) in mm}: The maximum interpenetration distance between the hand and all grasped objects.
    \item \textbf{Diversity (Div.) in radians}: Grasp diversity is determined by calculating the standard deviation of 
    $\matr{g}$ across all successful grasps.
    \item \textbf{Efficiency (Eff.) in seconds}: The computational time required to generate a batch of 256 grasps on an NVIDIA A100.
\end{enumerate}

\begin{table}[t]
    \centering
    \begin{adjustbox}{max width=0.9\linewidth}
         \begin{tabular}{lccccccc}
            \toprule
             & \multicolumn{2}{c}{\textbf{SR  $\uparrow$}} & \multicolumn{2}{c}{\textbf{Pene. $\downarrow$}} & \textbf{Eff. $\downarrow$} & \textbf{Div.} $\uparrow$  \\
            \cmidrule(l){2-3}  \cmidrule(l){4-5} \cmidrule(lr){6-6} \cmidrule(lr){7-7}
            \textbf{Method} & \textbf{\datasetshort{} } & \textbf{ \graspemshort{}} & \textbf{ \datasetshort{}} & \textbf{\graspemshort{}}  & \textbf{Avg.} & \textbf{Avg.} \\
            \midrule
            \multigraspshort{}-1 &\textbf{66.84} & \textbf{65.39} & \textbf{1.14} & \textbf{1.27} & 600 & 0.284\\
            \diffusionshort{}-1  &46.95 &45.23  & 5.55 & 5.59 & \textbf{0.8} & 0.321\\
            \methodshort{}-1     &50.04 & 40.78 & 1.73 & 2.16 & 900 & \textbf{0.332}\\    
            \midrule
            \multigraspshort{}-2 &\textbf{22.46} & 16.48 & 2.30 & 2.83 & 750 & 0.347\\
            \diffusionshort{}-2  &18.83   &      23.00  & 5.84 & 6.28 & \textbf{0.8} & 0.359\\
            \methodshort{}-2     &21.21 & \textbf{32.03} & \textbf{2.14} & \textbf{1.78} & 900 & \textbf{0.367}\\
            \midrule
            \multigraspshort{}-3 &10.78 & 3.55  & 3.39 & 4.04 &  900 & 0.340\\
            \diffusionshort{}-3  &11.05 & 9.22  & 5.93   &  6.47 & \textbf{0.8} & 0.334\\
            \methodshort{}-3     &\textbf{19.49} & \textbf{21.05} & \textbf{2.23} & \textbf{2.23} & 900 & \textbf{0.349}\\
            \midrule
            \multigraspshort{}-4 & 0.90  & 0.47  & 5.17 & 6.27 & 1000& \textbf{0.329}\\
            \diffusionshort{}-4  & \textbf{3.01}  & 1.68 & 6.18& 6.69 & \textbf{0.8} & 0.293\\
            \methodshort{}-4     & 2.93  & \textbf{5.04} & \textbf{2.70} & \textbf{2.62} & 900 & 0.312\\
            \bottomrule
        \end{tabular}
    \end{adjustbox}
    \caption{\textbf{Simulation results}. \multigraspshort{}, \methodshort{}, \diffusionshort{}, \graspemshort{}, and \datasetshort{} are short for the \multigrasp{}, \methodname{}, \diffusionname{}, \graspem{}, and \datasetname{}, respectively. The $-i$ following the method name denotes the number of objects used for grasp generation. $\uparrow$($ \downarrow$), the higher (lower), the better.}
    \label{tab:sim_sr_pen}
\end{table}

The quantitative results are presented in \tabref{tab:sim_sr_pen}, while \figref{fig:sim_visual_result} qualitatively illustrates a few grasps. The results demonstrate that \methodname{} achieves the highest success rate and the lowest penetration depth when grasping two or more objects.

\begin{figure}[t]
    \centering
    \includegraphics[width=0.9\linewidth]{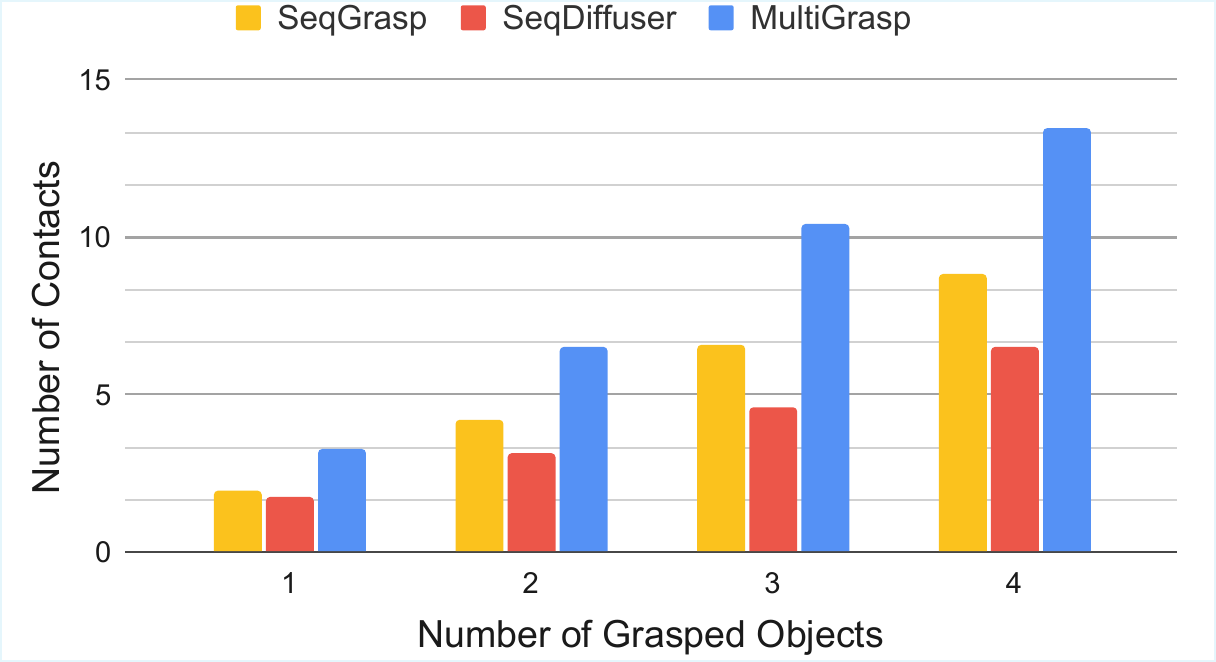}
\caption{\textbf{Number of hand-object contacts} for different methods.}
\label{fig:contact_chart}
\end{figure}

\setulcolor{black}
\setul{0.3ex}{0.2ex}
However, the lower performance of SeqGrasp on single-object grasps is attributed to the fewer contact points compared to MultiGrasp, as shown in \figref{fig:contact_chart}. For single-object grasps, SeqGrasp and SeqDiffuser establish approximately two contacts on average, whereas MultiGrasp achieves three or more due to it engaging all the available DoF. However, as the number of objects increases, the overall hand-object penetration increases for MultiGrasp but not for SeqGrasp.

We observe a notable decline in performance across all methods when transitioning from three-object to four-object grasps. We hypothesize that this decline occurs because the three previously grasped objects occupy substantial space within the Allegro Hand, pushing the fourth object grasp to the limits of the hand’s kinematic redundancy. Additionally, as the number of grasped objects increases, object-object interactions grow exponentially, making the task considerably more challenging, a finding also reported in \cite{multigrasp}. Nevertheless, \methodname{} demonstrates superior performance in scenarios involving three or more objects.

Another notable observation is that \diffusionname{} generates grasps with high penetration, which aligns with previous work on single-object diffusion-based grasp sampling~\cite{dexdiffuser}.
Still, \diffusionname{} is valuable because it generates grasps 750-1250 times faster than \methodname{} and \multigrasp{}.

\subsection{Real-World Experiments}

\begin{figure}[t]
    \centering
    \includegraphics[width=\linewidth]{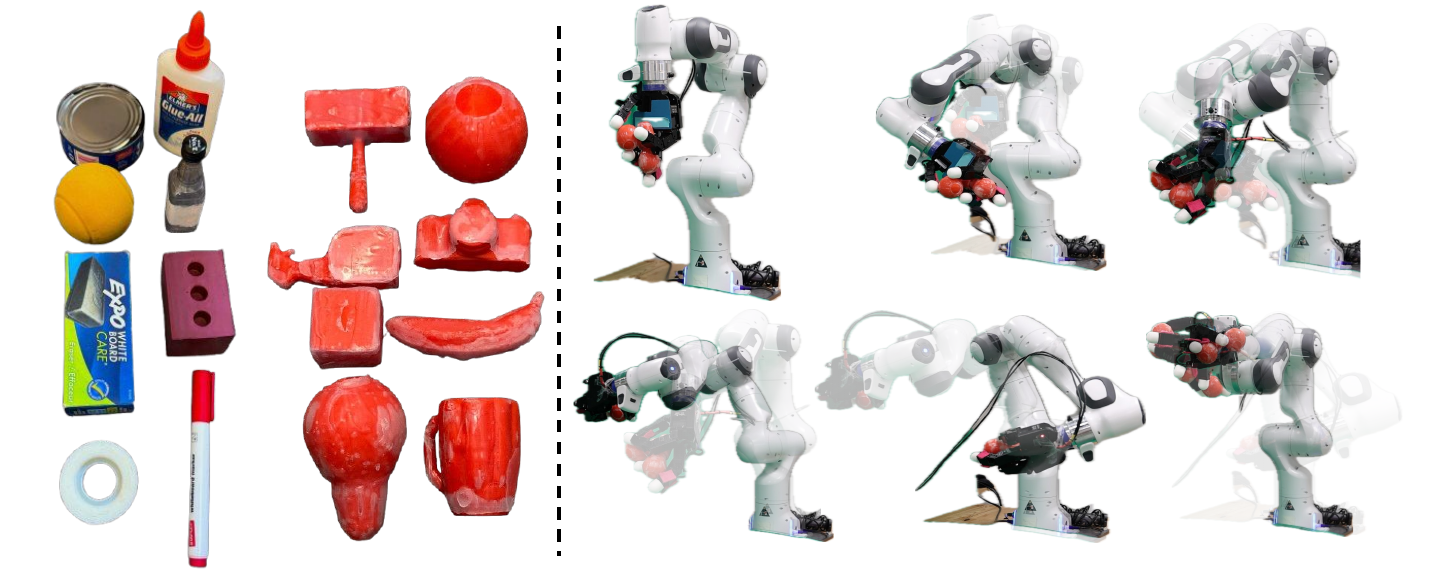}
\caption{Left: Two object sets for the grasp stability test. Right: \textbf{The dynamic stability test procedure.}}
\label{fig:objectsandtest}
\end{figure}

To evaluate grasp stability in the real world, we tested the generated grasps on a real Allegro Hand mounted to a \franka{} robotic arm. We evaluated the grasps on the two object sets visualized in \figref{fig:objectsandtest}: one comprising eight 3D printed \datasetname{} test objects, and another comprising eight everyday objects. To enhance surface friction, we coated the printed object with silicone gel and attached an anti-slip grip to the palm. 

\begin{figure}[b]
    \centering
     \begin{tikzpicture}
        \node[anchor=south west, inner sep=0] (image) at (0,0) {\includegraphics[width=0.95\linewidth]{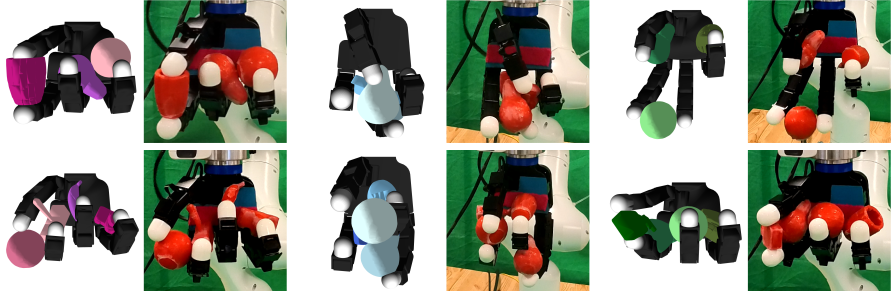}};
        \begin{scope}[x={(image.south east)}, y={(image.north west)}]  
            \node[black] at  (0.15,-0.1) {\methodname{}};
            \node[black] at (0.5,-0.1) {\multigrasp{}{}};
            \node[black] at (0.85,-0.1) {\diffusionname{}};
            
        \end{scope}
    \end{tikzpicture}
\caption{\textbf{Grasp replication on real hardware.} First row: three-object grasp, second row: four-object grasp.}
\label{fig:sim_real_comparison}
\end{figure}

We followed the procedure outlined in \cite{yao2023exploiting} to replicate the grasp on the real hardware. This procedure involves positioning each object as closely as possible with its generated pose and then closing the relevant joints of the hand to retain the objects, as exemplified in \figref{fig:sim_real_comparison}. For grasps with severe penetration issues, we placed the object in the closest feasible real-world position. For \multigrasp{} and \methodname{}, we replicated the successful grasps with the lowest energy, while for \diffusionname{}, we randomly chose one of the generated grasps. Due to the larger real-world objects, four-object grasping exceeded the workspace limits of the Allegro Hand. Thus, we only evaluated three-object grasps on that object set. In contrast, both three- and four-object grasping were tested on the 3D-printed object set.

We evaluated the grasps with the grasp stability test shown in \figref{fig:objectsandtest}. In this test, the arm followed a randomly generated dynamic trajectory executed with the highest achievable acceleration. The arm was then moved to a holding position with the palm facing downward. From this position, the last joint rotated by $\pm$90\textdegree{}. This test comprehensively assessed the grasp stability by evaluating both translational and rotational disturbances.

\begin{table}[t]
    \centering
    \begin{adjustbox}{max width=\linewidth}
    \begin{tabular}{cccccccc}
        \toprule
        Real exp. & & \methodshort{}-3 & \methodshort{}-4 & \diffusionshort{}-3 & \diffusionshort{}-4 & \multigraspshort{}-3 & \multigraspshort{}-4  \\
        \midrule
        \multirow{2}{*}{Printed Obj.} & Succ. trials & 5/10 & 5/10 & 4/10 & 0/10 & 1/10 & 1/7 \\
        & Succ. objects & 2.3$\pm$0.9/3 & 3.4$\pm$0.8/4 & 2.1$\pm$0.9/3 & 1.5$\pm$0.8/4 & 1.2$\pm$1.1/3 & 1.6$\pm$1.5/4 \\
        \midrule
        \multirow{2}{*}{Real Obj.} & Succ. trials & 5/10 & / & 0/10 & / & 0/6 & / \\
        & Succ. objects & 2.1$\pm$1.2/3 & / & 1.8$\pm$0.4/3 & / & 1.0$\pm$0.6/3 & / \\
        \bottomrule
    \end{tabular}
    \end{adjustbox}
    \caption{\textbf{Real-world experimental results.} Each entry reports the number of successful trials out of 10 and the average number of objects successfully grasped in mean$\pm$std after the test.}
    \label{tab:realexp}
\end{table}

The experimental results are presented in \tabref{tab:realexp}. Similar to the simulation results, \methodname{} outperformed \diffusionname{} and \multigrasp{} in both three-object and four-object grasping tasks, achieving an average success rate of 50\%. The low success rate of \diffusionname{} (13\%) is primarily due to substantial object inter-penetration in the generated grasps, which meant that the closest physically feasible grasp replicable on the real hand differed substantially from the intended configuration. \multigrasp{} performed worst, with an average success rate of 8.7\%. The reason is its reliance on object-object contacts for grasp stability, which are sensitive to minor object displacements. This resulted in frequent failures, particularly in four-object grasping trials where \multigrasp{} achieved no successful trials and failed to generate valid grasps in three of ten attempts (last column in \tabref{tab:realexp}). Accounting for these additional failed trials lowers the average success rate for \multigrasp{} from 8.7\% to 6.7\%.

The real-world experiments highlight that our sequential grasping strategy enables robust and effective dexterous multi-object manipulation. By generating grasps iteratively and without requiring spatial proximity between objects, \methodname{} and \diffusionname{} mitigate the challenges posed by object interactions, making it a more reliable and practical approach for real-world applications. 

\section{Limitations and Future work}

Our work has two main limitations. First of all, neither SeqGrasp nor SeqDiffuser accounts for object sequences as OSes and object orders are uniformly sampled. The random sampling results in a diverse dataset that is useful for training and benchmarking, rather than the most optimal grasp sequence. Learning the optimal grasp sequences from human preferences is an interesting future work direction.

The second limitation is that the grasp generation process does not account for motion planning to the grasp pose. Consequently, it proved extremely challenging to find a collision-free path to the intended grasp pose, which is the primary reason no such experiments were conducted. However, we believe this limitation can be resolved by learning a policy conditioned on the final grasp pose using, \eg{}reinforcement learning.

\section{Conclusion}

We proposed \methodname{}, an algorithm for sequentially grasping multiple objects with a dexterous hand. \methodname{} combines \acp{os} and differentiable-force closure to generate stable grasps that maximize the hand's remaining \ac{dof} after each grasp. Using \methodname{}, we constructed \datasetname{}, currently the largest sequential grasping dataset, comprising 870K validated grasps across 509 diverse objects. This dataset enabled the training of \diffusionname{}, our diffusion-based sequential multi-object grasp sampler. The experimental evaluations demonstrated that \methodname{} and \diffusionname{} outperformed the simultaneous multi-object grasping baseline \multigrasp{}, achieving an 8.71\%-43.33\% higher average success rate. Moreover, \diffusionname{} proved to be 750-1250 times faster at generating grasps than \methodname{} and \multigrasp. In conclusion, this work demonstrates a fast and stable solution for generating sequential multi-object grasps, which we hope will pave the way for further research in multi-object grasping.
\balance
\bibliographystyle{IEEEtran}
\bibliography{ref}

\end{document}